\documentclass{article}
\PassOptionsToPackage{hyphens}{url}  
\usepackage{iclr2027_conference,times}
  \usepackage[T1]{fontenc}
  \usepackage[utf8]{inputenc}

\usepackage{amsmath,amsfonts,bm}

\def\eqref#1{equation~\ref{#1}}

\def\1{\bm{1}}

\DeclareMathAlphabet{\mathsfit}{\encodingdefault}{\sfdefault}{m}{sl}
\SetMathAlphabet{\mathsfit}{bold}{\encodingdefault}{\sfdefault}{bx}{n}

\newcommand{\demo}{\mathcal{D}}
\newcommand{\tutor}{\mathcal{T}}
\newcommand{\student}{\mathcal{M}}
\usepackage{hyperref}
\usepackage{url}
\usepackage{graphicx}
\usepackage{wrapfig}
\usepackage{booktabs}
\usepackage{xcolor}
\usepackage{amsmath,amssymb}
\usepackage{algorithm}
\usepackage[noend]{algpseudocode}
\usepackage{multirow}
\usepackage{array,calc,tabularx}
\usepackage{arydshln}
\usepackage[most]{tcolorbox}
\definecolor{promptink}{HTML}{2E2A26}
\definecolor{promptfill}{HTML}{FBF6EE}
\definecolor{amberline}{HTML}{B35E1C}
\definecolor{amberfill}{HTML}{F8DFBC}
\definecolor{demoline}{HTML}{8E7EA0}
\definecolor{demofill}{HTML}{EDE6F0}
\tcbset{promptbase/.style={enhanced, breakable, colback=promptfill, coltext=promptink, boxrule=0.6pt, arc=1.5pt, left=4pt, right=4pt, top=3pt, bottom=3pt, fonttitle=\small\bfseries, listing only,
  listing options={basicstyle=\ttfamily\scriptsize, breaklines=true, breakatwhitespace=true, columns=fullflexible, keepspaces=true, upquote=true, showstringspaces=false}},
  prompttutor/.style={promptbase, colframe=amberline, colbacktitle=amberfill, coltitle=promptink},
  promptdemo/.style={promptbase, colframe=demoline, colbacktitle=demofill, coltitle=promptink}}
\newcommand{\promptbox}[3]{\tcbinputlisting{#1, title={#2}, listing file={#3}}}

\iclrfinalcopy
\hypersetup{colorlinks=true, linkcolor={blue!60!black}, citecolor={blue!60!black}, urlcolor={blue!60!black}}

\title{Splicing Hints into Reasoning Traces: \\ Learning from Demonstrator Failures}

\newcommand{\authorbox}[3]{\begin{minipage}[t]{0.44\textwidth}\raggedright
  \textbf{#1}\\ \normalfont #2\\ \texttt{#3}\end{minipage}}
\author{%
\authorbox{Hamed Mahdavi}{Pennsylvania State University}{hmm5832@psu.edu}\hspace{0.04\textwidth}%
\authorbox{Pouria Mahdavinia}{Pennsylvania State University}{pmahdavi@psu.edu}\\[3.6em]
\authorbox{Majid Farhadi}{Google}{mfarhadi@google.com}\hspace{0.04\textwidth}%
\authorbox{Vasant Honavar}{Pennsylvania State University}{vuh14@psu.edu}%
}

\begin{document}
\maketitle
\lhead{Preprint}

\begin{abstract}
Reinforcement learning with verifiable rewards improves mathematical
reasoning when a final answer can be checked, but a correct answer does
not certify the derivation behind it, and proof problems offer no answer
to check. Replacing the check with a reasoning model as an in-loop
verifier faces a fundamental obstacle: the policy must already produce occasional
correct attempts for any reward to exist, which it may never do at
olympiad level. We instead move verification and repair out of the training
loop and into offline data generation with SPLICE,
a workflow in which a second model, given only a reference solution,
edits a reasoning model's traces as they are generated: it
verifies progress, truncates a failed trace at its last verified step,
and splices in an adaptively strengthened hint until progress resumes.
The resulting dataset includes repaired examples whose sampled unaided
trajectory failed, with failed continuations removed and
progress repeatedly checked against the reference solution. A model
finetuned on SPLICE traces outperforms the same model distilled from
the reasoning model's unedited traces on AIME and, under a grading
protocol validated against human grades, on olympiad proofs. The
trained model's own traces show a corresponding behavioral signature:
at difficult points it announces a branch in the style of the spliced
hints and takes riskier steps, with predictive entropy rising there and
just after, a pattern absent from the model trained on unedited traces.

\end{abstract}

\section{Introduction}
\label{sec:intro}

Reinforcement learning with verifiable rewards (RLVR) improves mathematical reasoning when final answers can be checked automatically
\citep{deepseekai2026deepseekr1incentivizingreasoningcapability,
shao2024deepseekmathpushinglimitsmathematical,
kimiteam2025kimik15scalingreinforcement,
lambert2025tulu3pushingfrontiers}. Proof problems are different: the solution is the argument itself, and a correct final answer need not certify correct reasoning
\citep{mahdavi2025brains,petrov2025prooforbluff}. A natural alternative is a strong reasoning model acting as a verifier
\citep{zhang2025generativeverifiersrewardmodeling,bai2022constitutionalaiharmlessnessai,
pmlr-v235-lee24t,pmlr-v235-yuan24d}. With a reference solution, such verifiers can agree closely with human graders
\citep{luong2025towardsrobustmath,ma2025reliablefinegrained}, but two obstacles remain. First, reinforcement learning needs occasional correct attempts to observe reward, which may be absent on olympiad-level problems
\citep{qu2026popelearningreasonhard,zhang2026surveyhintbasedrlvr,guo2026latentreviselearningzerohit}. Second, querying a verifier on every rollout compounds the dominant cost of large-scale RL
\citep{sheng2025hybridflow}.

Test-time scaffolds decompose reasoning into generation, verification, and revision
\citep{NEURIPS2023_91edff07,NEURIPS2023_1b44b878,NEURIPS2023_271db992}. They can reach very high performance when the underlying model is strong
\citep{huang2025gemini25imo,wei-openai-imo-2025-proofs}, but they depend on the model finding and fixing its own errors, which weaker models cannot do without external feedback
\citep{huang2024largelanguagemodelsselfcorrect}. Existing proof-training recipes likewise generally begin with a model that already writes proofs and train on its traces
\citep{hamid2026spirallearningsearchaggregate,qu2026qednanoteachingtinymodel,
deepseekai2025deepseekmathv2}. We instead supply verification and repair externally, only during data generation.

Our target is the single-attempt proof generation of an instruction-tuned \textbf{student} not post-trained for extended reasoning. A standard baseline finetunes it on traces from a larger \textbf{demonstrator}
\citep{yang2025qwen3technicalreport}, either directly as distillation
\citep{hinton2015distillingknowledgeneuralnetwork,ho-etal-2023-large,
magister-etal-2023-teaching,hsieh-etal-2023-distilling} or after rejection sampling
\citep{zelikman2022starbootstrappingreasoningreasoning,
dong2023raftrewardrankedfinetuning,gulcehre2023reinforcedselftrainingrestlanguage,
singh2024humandatascalingselftraining,yuan2023scalingrelationshiplearningmathematical}.
Both learn only from what the demonstrator can produce unaided.

\begin{wrapfigure}{r}{0.55\linewidth}
\vspace{-\baselineskip}
\centering
\includegraphics[width=\linewidth]{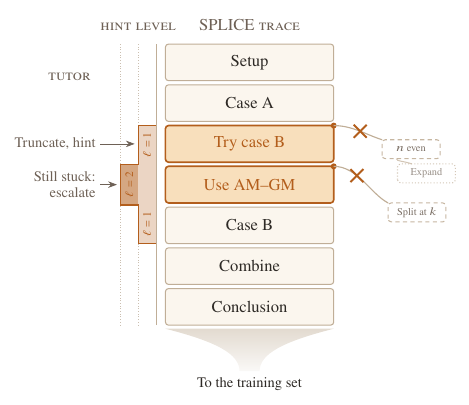}
\caption{One SPLICE trace. When the tutor finds no newly verified step, the failed continuation is discarded, an adaptively strengthened hint is rewritten in the trace's style and appended to the last constructive prefix, and generation resumes. Hint level rises after stalls and falls after progress; only the surviving trace is retained.}
\label{fig:1}
\end{wrapfigure}

We propose \textbf{SPLICE}, which repairs a demonstrator's failures as they occur (Figure~\ref{fig:1}). A third model, the \textbf{tutor}, receives a reference solution, checks whether the trace has advanced, and writes a hint only when needed; it is never asked to solve the problem. Because checking against a reference can be substantially easier than solving
\citep{luong2025towardsrobustmath,ma2025reliablefinegrained}, the tutor can contribute competence without supplying a full reasoning trace and is absent at test time. SPLICE truncates failed continuations at the last constructive step, adaptively strengthens a hint until progress is restored in the demonstrator's own style, and resumes generation. A verifier applied after the attempt could only have scored the failed trace; acting during generation, the tutor changes what the demonstrator can reach. Thus the dataset can contain completed solutions built from sampled unaided trajectories that failed, while preserving the demonstrator's successful decisions. A failure here is a trajectory that stalled, not a problem the demonstrator could never solve.

The intervention is deliberately asymmetric. Successful reasoning remains untouched; an incorrect or stalled continuation is discarded rather than revised in place; and hint strength rises only after repeated stalls and falls when progress resumes. This makes the resulting target a single coherent solution rather than a tutor--student dialogue. It also gives each trace a generation record---where progress stopped, how much help was required, and whether the demonstrator recovered---that can later be used to diagnose harmful training examples.

We use Qwen3-4B-Instruct-2507 as student, Qwen3-30B-A3B-Thinking as demonstrator
\citep{yang2025qwen3technicalreport}, and Gemini 2.5 Pro as tutor
\citep{comanici2025gemini25}, with cheaper family members for lighter stages (Appendix~\ref{app:tutor-models}). The controlled baseline finetunes the same student on untutored traces from the same demonstrator
\citep{ho-etal-2023-large,magister-etal-2023-teaching,
muennighoff2025s1simpletesttimescaling,ye2025limoreasoning}. Under the human-validated grading protocol of \citet{mahdavi2025brains}, filtered SPLICE training raises the mean proof score from 0.97 to 1.15, an 18\% gain, improving on 33 of 45 problems and worsening on 12, and doubles the number of attempts graded 6/7 or above; AIME accuracy rises from 60.6\% to 64.4\%. The trained student's traces also show a distinctive learned behavior: at difficult points it emits self-hint-like branch cues and predictive entropy rises in and especially after those passages.

\paragraph{Relation to prior work.}
SPLICE differs from outcome-level RLVR and rejection sampling, which learn from
successful completed rollouts: when an unaided trajectory stalls, SPLICE edits
the trajectory itself and supplies information needed to resume reasoning
\citep{qu2026popelearningreasonhard,zhang2026surveyhintbasedrlvr,
guo2026latentreviselearningzerohit}. It is closest to hint-based methods such as
POPE, ADHint, and HiLL, but those inject guidance into the learner's RL
rollouts, whereas SPLICE spends privileged information once, during offline
demonstrator trace construction, and the student receives no hinted prompt at
test time \citep{qu2026popelearningreasonhard,zhang2025adhintadaptivehints,
xia2026learninghintreinforcementlearning}. Finally, unlike ordinary or
solution-conditioned teacher distillation, the reference-conditioned tutor
does not author the surrounding solution: it identifies a stalled prefix and
intervenes locally, while the demonstrator remains the source of the reasoning
trajectory \citep{kaur2026rethinkingonpolicyselfdistillation,
zhang2026anchoredresidualprivilegeddistillation}. Appendix~\ref{app:related}
gives the fuller comparison.

Our contributions are: \textbf{(i)} an offline trace-editing workflow that turns demonstrator failures into supervised training signal; \textbf{(ii)} a controlled comparison against untutored distillation, together with a rejection-sampled-success baseline; \textbf{(iii)} an analysis showing that SPLICE-trained students learn where to branch and explore; and \textbf{(iv)} a generation-record-based taxonomy of failure modes whose removal produces distinct, diagnostically consistent changes in evaluation behavior.

\section{SPLICE: editing reasoning traces during generation}
\label{sec:method}

SPLICE generates supervised reasoning traces by controlling the demonstrator's decoding rather than collecting its finished traces: generation can be stopped, the trace cut back to any prefix, and decoding resumed from there. We take efficient trace distillation as the baseline \citep{muennighoff2025s1simpletesttimescaling}; the tutor, holding a reference solution, watches the trace as
the demonstrator writes it; where progress stalls, the trace is cut back
to its last verified step, a hint is spliced in, and the demonstrator
resumes from there, until the problem is solved. The trace that survives
is one SPLICE trace.

\subsection{Building blocks}
\label{sec:notation}

\begin{wrapfigure}{r}{0.55\linewidth}
\vspace{-\baselineskip}
\centering
\includegraphics[width=\linewidth]{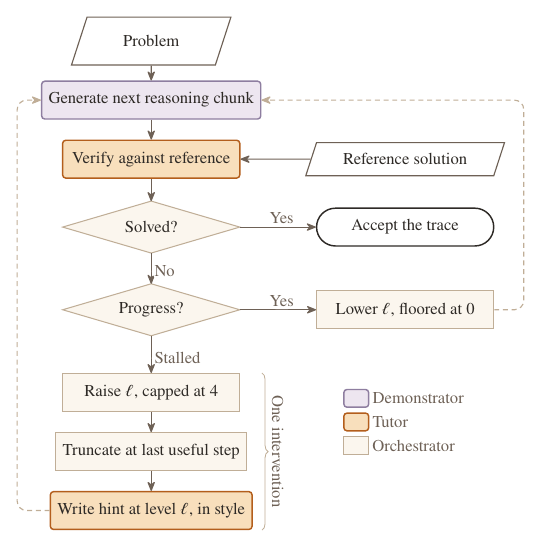}
\caption{One iteration of the SPLICE loop. Colors mark which model performs each step; the bracketed steps are one intervention.}
\label{fig:2}
\end{wrapfigure}
We write \(\demo\) for the demonstrator, \(\tutor\) for the tutor, and \(\student\) for
the student. In a loop, the \textbf{demonstrator} \(\demo\) writes the
reasoning trace, chunk by chunk. We assume only that its reasoning traces are
accessible and that generation can be resumed from a prefix we supply,
which open-weight models provide. The \textbf{tutor} \(\tutor\) is the
model queried during generation; we assume only that it can compare a
written solution against a reference solution. The \textbf{student}
\(\student\) is the instruction-tuned model finetuned on the finished
dataset; it takes no part in generation. An instance is a pair \((p, r)\) of a problem \(p\) and a
human-written reference solution \(r\). The demonstrator's output splits
into the reasoning trace \(c = (c_1,\dots,c_n)\), whose steps \(c_i\) are the
paragraphs, separated by blank lines, and the final solution \(a\), the
text following the reasoning block. Generation proceeds in iterations
\(t = 1, \dots, B\); iteration \(t\) produces one chunk, a single call of at
most \(L\) tokens.

Figure~\ref{fig:2} shows one iteration of the SPLICE loop. Each stage is one prompt to one
model, and we denote each with a function notation. Generation is the single call \(\textsc{Gen}\), which
extends the trace by one chunk; every other stage belongs to the tutor.
\(\textsc{Decompose}\) reads \(r\) once and returns the ordered substeps
\(S\). \(\textsc{Accept}\) judges a finished solution \(a\) against \(r\) and \(S\),
returning a verdict together with a list \(E\) of errors located in \(a\). \(\textsc{Assess}\) grades each substep against the solution and
finds the last paragraph of the trace that did constructive work: a
paragraph is constructive if it derives a correct substep, sets a
relevant subgoal, validates a claim or identifies an error, proposes a
new angle, takes up an earlier hint, or repairs an earlier mistake, and
not if it merely restates what was done, repeats a known error, stalls
without proposing an action, or jumps to a conclusion.
\(\textsc{Hint}\) writes the content of an intervention at the current
level, and \(\textsc{Restyle}\) rewrites that content in the
demonstrator's idiom, using patterns \(\Pi\) drawn from the
demonstrator's own reasoning and the retained prefix. Two further stages keep generation inside its
budget: \(\textsc{Shorten}\) compresses an over-long trace, and
\(\textsc{Stopped}\) reports whether a finished solution is already
sitting inside the reasoning block. Only \(\textsc{Gen}\) lacks access to \(r\), and only \(\textsc{Accept}\) receives \(a\) without \(c\). A deterministic orchestrator runs the loop, applies truncations, and stores state; it makes no semantic judgments.

Three quantities drive the loop, all read off \(\textsc{Assess}\). The
status map \(\sigma_t\) assigns each substep \(s_i\) of \(S\) a status in
\(\{\textit{correct}, \textit{incorrect}, \textit{missing}\}\) according to
\(a\). Let \(v_t = |\{\, i : \sigma_t(i) = \textit{correct} \,\}|\) be the
number of substeps marked \textit{correct} at iteration \(t\), and \(V_t = \max_{s \le t}
v_s\) the best count reached so far. Iteration \(t\) is a \textbf{stall}
if it does not raise the best count, that is, \(v_t \le V_{t-1}\);
otherwise it is \textbf{progress}. The \textbf{pointer} \(k_t\) is the
number of consecutive substeps marked \textit{correct} from the start,
the largest \(k\) with \(\sigma_t(i) = \textit{correct}\) for all
\(i \le k\). The \textbf{cut index} \(u_t\) is the last paragraph of
\(c\) that \(\textsc{Assess}\) marks constructive.

The \textbf{hint level} \(\ell_t \in \{0,1,2,3,4\}\), initialised to \(0\),
controls how much a hint may reveal, and the levels form a ladder: level
\(0\) names the next mathematical goal in one line, with no method; \(1\)
adds the specific claim or constraint to establish next; \(2\) gives a
concrete method for that claim, which identity or inequality to apply
and what to check while applying it; \(3\) outlines the whole corrective
argument with its intermediate claims; and \(4\) carries out the single
step the demonstrator is stuck on, and nothing beyond it.

\subsection{The SPLICE loop}
\label{sec:loop}

\begin{algorithm}[t]
\small
\algrenewcommand{\algorithmiccomment}[1]{\hfill$\triangleright$ {\footnotesize #1}}
\caption{SPLICE: generating one reasoning trace. The problem $p$, reference solution $r$, and decomposition $S$ are shared by all procedures.}
\label{alg:splice-loop}
\begin{algorithmic}
\Require problem $p$, reference solution $r$; constants $B$, $L$, $N$ (Appendix~\ref{app:constants})
\Ensure an accepted trace $(p, c, a)$, or failure
\State $S \gets \textsc{Decompose}_{\tutor}(p, r)$; \; $c \gets (\,)$; \; $a \gets \varepsilon$; \; $\ell \gets 0$; \; $V \gets 0$
\For{$t = 1$ \textbf{to} $B$}
    \State $(c, a, \mathit{done}) \gets \Call{Extend}{c, a}$ \Comment{demonstrator writes one chunk}
    \If{\textbf{not} $\mathit{done}$} \textbf{continue}
    \EndIf
    \State $(\mathit{ok}, E) \gets \textsc{Accept}_{\tutor}(p, r, S, a)$ \Comment{$N$ samples, all must agree}
    \If{$\mathit{ok}$} \Return $(p, c, a)$ \Comment{accepted: enters the dataset}
    \EndIf
    \State $(c, a, \ell, V) \gets \Call{Intervene}{c, a, E, \ell, V}$ \Comment{tutor cuts and hints}
\EndFor
\State \Return failure \Comment{iteration budget $B$ exhausted}
\end{algorithmic}
\vspace{2pt}\hrule\vspace{2pt}
\noindent
\begin{minipage}[t]{0.49\linewidth}
\begin{algorithmic}
\Procedure{Extend}{$c, a$}
    \State $x \gets \textsc{Gen}_{\demo}(p, c, a)$ \Comment{at most $L$ tokens}
    \If{the reasoning block is still open}
        \If{\Call{Repeats}{$x$}}
            \State cut $x$ at the repeat; close the block
        \ElsIf{$\textsc{Stopped}_{\tutor}(p, c + x)$}
            \State $(c, a) \gets \Call{SplitAtSolution}{c + x}$
        \Else
            \State \Return $(c + \Call{Paragraphs}{x},\ a,\ \mathit{false})$
        \EndIf
    \EndIf
    \State $a \gets a + x$
    \If{$x$ hit the cap $L$} \Return $(c, a, \mathit{false})$
    \EndIf
    \State \Return $(c, a, \mathit{true})$ \Comment{a solution exists}
\EndProcedure
\end{algorithmic}
\end{minipage}\hfill
\begin{minipage}[t]{0.49\linewidth}
\begin{algorithmic}
\Procedure{Intervene}{$c, a, E, \ell, V$}
    \State $(\sigma, k, u) \gets \textsc{Assess}_{\tutor}(p, S, c, a, E)$
    \State $v \gets |\{\, i : \sigma(i) = \textit{correct} \,\}|$
    \If{$v > V$} $\ell \gets \max(0,\ \ell - 1)$ \Comment{progress}
    \Else \ $\ell \gets \min(4,\ \ell + 1)$ \Comment{stall}
    \EndIf
    \State $V \gets \max(V,\ v)$
    \State $c \gets c_{1:u}$ \Comment{truncate}
    \State $h_0 \gets \textsc{Hint}_{\tutor}(p, r, S, c, a, \sigma, k, \ell, E)$
    \State $h \gets \textsc{Restyle}_{\tutor}(p, h_0, c, \Pi)$
    \State \Return $(c + h,\ \varepsilon,\ \ell,\ V)$ \Comment{resume reasoning}
\EndProcedure
\end{algorithmic}
\end{minipage}
\end{algorithm}

Algorithm~\ref{alg:splice-loop} alternates generation with assessment. \textsc{Extend} handles unfinished reasoning: \(\textsc{Repeats}\) cuts repeated passages, \(\textsc{Stopped}\) detects a prematurely closed reasoning block, \(\textsc{SplitAtSolution}\) separates an embedded solution, and \(\textsc{Paragraphs}\) forms steps. If \(c\) exceeds \(L_{\max}\), \(\textsc{Shorten}\) compresses it. These are budget controls rather than the core intervention; Section~\ref{sec:failure-modes} measures the harm caused by shortening.

Once a solution exists, the main loop decides whether to keep the
trace: \(\textsc{Accept}\) is sampled \(N\) times and every attempt must
agree, since one judgment from a strong model is not reliable enough to
gate a training example; the grading protocol of
Section~\ref{sec:protocol} repeats samples for the same reason.
Because \(\textsc{Accept}\) sees \(a\) without \(c\), a trace cannot be
accepted on the strength of reasoning its written solution does not
carry. A trace accepted at \(t = 1\) is one the demonstrator solved unaided, with no
hint spliced in; we flag these traces, since they form the Untutored-Correct
control of Section~\ref{sec:stores}.

If the trace is rejected, \textsc{Intervene} carries out the
intervention, which is where the method differs from resampling. The
hint level moves first: it falls
on progress and rises on a stall, so the amount revealed tracks what
this demonstrator has needed on this attempt rather than what the
problem is presumed to require. Because \(\ell\) is capped at \(4\), the pipeline can fail: a problem the
demonstrator cannot advance on even when handed the step is never
completed. At each intervention the trace is cut back to \(c_{1:u}\), discarding
everything after the last paragraph \textsc{Assess} marked constructive.
The discarded text never reaches the dataset, and the demonstrator is
never asked to revise it: the hint is appended to the constructive prefix
and generation resumes from there.

The hint targets \(s_{k+1}\), the first substep not marked \textit{correct}. \(\textsc{Hint}\) fixes its mathematical content at level \(\ell\), while \(\textsc{Restyle}\) separately rewrites it in the demonstrator's idiom so the supervised target does not alternate visibly between speakers. Each trace stores hint levels, interventions, inserted text, and statuses \(\sigma_t\), enabling the analyses in Sections~\ref{sec:failure-modes} and~\ref{sec:stores}.

Two design choices are central. First, progress is measured against the best substep count reached so far rather than the immediately preceding iteration, because truncation, stochastic tutor judgments, and a demonstrator that ignores the hint and drifts off course all make the raw count nonmonotone. Second, failed continuations are removed before a hint is inserted. Each SPLICE trace therefore contains the constructive prefix, the intervention, and the continuation that follows, never the demonstrator's known error and its subsequent correction. This is what distinguishes trace editing from ordinary critique-and-revision data.

\subsection{Making the demonstrator follow a hint}
\label{sec:following}

A reasoning model does not take instruction from inside its own trace.
Text inserted into a chain of thought is often ignored, even when it is
the correct solution \citep{cuestaramirez2025notthinkingstraight}, and an
insertion can instead make the model abandon its reasoning and close the
block early \citep{tang2025concisehint,wu2025interruptible}. We saw both
in our early runs: resumed from a prefix that ends in a hint, the
demonstrator would either close the reasoning block at once, or open the
next paragraph with a pivot (``Wait'', ``But'', ``Alternatively'',
``Actually'') that discarded the hint and returned to its own line. Three
measures address this. Restyling (Section~\ref{sec:loop}) makes the hint
read as the demonstrator's own thought. Every hint opens by refusing
closure: a statement that the reasoning is not finished, one verification
to perform first, and two to five explicit questions drawn from the errors
\(E\) and the next substep \(s_{k+1}\), which restyling keeps as
questions (a level-\(4\) hint replaces the questions with an instruction
to check the supplied derivation and integrate it). For the first \(200\)
tokens after an insertion, we suppress two kinds of tokens at decoding:
the end-of-reasoning token, so the block cannot close, and a fixed list of
pivot words (Appendix~\ref{app:constants}), so the demonstrator has to
engage with the hint before it can reject it. These measures raise the
rate at which a hint is taken up but do not make it certain; the problems
on which every hint is ignored are the ones the loop fails on.

\subsection{Problem set}
\label{sec:training-data}

We ran the loop on 2,500 problems: AIME problems from 1983 to 2023, and
proof problems from the IMO (1959 to 2024), the IMO Shortlist (1966 to
2023) and Longlists, Shortlist-based team selection tests, Baltic Way,
Tuymaada, USAMO, and the Balkan MO Shortlist, none later than 2024, with
reference solutions from the corresponding Art of Problem Solving (AoPS)
threads. The pool is disjoint from the evaluation exams of
Section~\ref{sec:datasets}: it contains no AIME problem after 2023, no
Canadian olympiad problem, no Iranian olympiad problem after 2021, and no
IMO problem after 2024. It accepted traces for 2,130 of them, 778 AIME and 1,352 proof problems; on the rest the budget ran out
at the top hint level with the demonstrator ignoring or contradicting
every hint, so the dataset is selected on completion by construction.

\section{The SPLICE traces}
\label{sec:produces}

\subsection{Yield and coverage}
\label{sec:yield}

\begin{figure}[t]
\centering
\includegraphics[width=\linewidth]{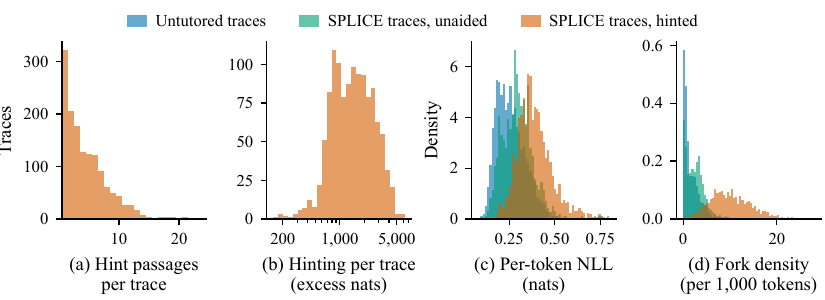}
\caption{SPLICE trace statistics: (a) hint passages per hinted trace; (b) amount of hinting per trace, measured by excess hint-token negative log-likelihood; (c) per-token negative log-likelihood and (d) fork density under the demonstrator's base model for untutored, unaided SPLICE, and hinted SPLICE traces.}
\label{fig:3}
\end{figure}

Of 2,130 accepted traces, 768 are unaided and 1,362 contain at least one spliced hint. Hinted traces are markedly less probable under the demonstrator's base model than untutored traces: median per-token negative log-likelihood is 0.25 nats for untutored traces, 0.29 for unaided SPLICE traces, and 0.38 for hinted traces; fork density (tokens per thousand with predictive entropy above 2 nats) is 1.0, 2.4, and 9.9, respectively (Figure~\ref{fig:3}). Hinting is long-tailed, and its upper tail contains the excessive-hinting failure mode analyzed below.

\subsection{Failure modes of the pipeline}
\label{sec:failure-modes}

The generation record exposes three systematic defects among the 2,130 accepted traces: compression artifacts from \(\textsc{Shorten}\) (98 traces), leakage of the reasoning block into the final solution (18), and excessive hinting at the maximum level without verified progress (238). The first two corrupt the generated target; the third records failure to progress and can teach that behavior. We therefore predicted that removing the first two would improve the top of the score distribution, while removing excessive hinting would primarily improve termination; Section~\ref{sec:filtering} tests these predictions. The same cut also drops 10 traces whose write-up answers a different problem from the one posed.

\section{Results}
\label{sec:results}

\subsection{Datasets and evaluation protocol}
\label{sec:protocol}
\label{sec:datasets}

We finetune the same student, Qwen3-4B-Instruct-2507, with one recipe on
five datasets, so that
the training data are the only experimental variable.
\textbf{Untutored} is the distillation baseline: the demonstrator's
2,500 traces, generated with no tutor. \textbf{Untutored-Correct} is the
rejection-sampling baseline: the 768 of those traces that solved the
problem unaided; containing no hint, it is also the negative control for the self-hint
analysis of Section~\ref{sec:stores}. \textbf{SPLICE-NF} is all 2,130 accepted
SPLICE traces; \textbf{SPLICE} removes the generation-failure
traces (2,009); \textbf{SPLICE-HC} also removes excessive hinting
(1,790). The first two against SPLICE ask whether repairing the
demonstrator's failures outperforms distilling or discarding them; the
three SPLICE datasets isolate each filter. We evaluate each trained student single-attempt, with no scaffold,
retries or tutor, on two tracks. The AIME track is the 90 problems of
AIME 2024--2026, eight samples each, scored by the boxed answer. The
proof track is 45 problems from eight Canadian, Iranian and IMO exams of
2024--2026, 32 samples each, graded on the final write-up and never on
the reasoning trace.

We grade with the protocol of \citet{mahdavi2025brains}: each sample
is matched to a cluster of the reference solutions and scored from 0 to
7 against a rubric authored for that cluster. Scores aggregate three judges, one over repeated draws; the equal-weighted
aggregate agrees with human grades at Pearson \(r=0.80\) on 90 human-graded
proofs. Appendix~\ref{app:grading} gives the judges, the weighting, the
validation and the full scoring rule; thresholds below apply to the
aggregate score. Beyond the mean score, we report two threshold metrics
for proofs: existence at \(T\), the share of the 45 problems with at
least one of the 32 attempts scoring \(\geq T\), and the number of
attempts scoring \(\geq T\) among the 1,440. Existence asks whether a
problem is within the student's reach at all; the count asks how often
it gets there. Intervals are 95\% percentile
bootstrap intervals over problems (45 for proofs, 90 for AIME), paired
across students, with 10,000 resamples; best-of-\(k\) intervals also
resample the 32 solutions within each problem.

\textbf{Force-closing.} Students finetuned on a few thousand traces sometimes fall into repetition and never emit the end-of-reasoning marker. On AIME this is harmless: we score the last boxed answer wherever it appears, so a looping attempt is scored like any other and is usually wrong. On proofs it leaves nothing to grade, since only the solution written after the reasoning block is graded, never the trace, and dropping such attempts would favor whichever student terminates most often. We therefore cut repetitions at their onset (or otherwise at the last complete paragraph), append a fixed closure, and allow at most 5,000 additional tokens.

\subsection{Performance of the trained students}
\label{sec:aime}
\label{sec:proof}

\begin{table}[t]
\caption{Student performance. AIME accuracy (\%), eight samples per problem, 30 per exam; proof grading over 1,440 attempts per student: aggregate mean score, existence, the share of the 45 problems with a sample scoring $\geq T$, and the number of attempts scoring $\geq T$. Best in bold, second in blue.}
\label{tab:results-2}
\centering
\footnotesize\setlength{\tabcolsep}{5pt}
\begin{tabular}{@{}lcccc:ccccc@{}}
\toprule
 & \multicolumn{4}{c:}{AIME accuracy (\%)} & \multicolumn{5}{c}{Proofs} \\
\cmidrule(lr){2-5}\cmidrule(lr){6-10}
 & \multicolumn{4}{c:}{} & & \multicolumn{2}{c}{Existence, $T$ (\%)} & \multicolumn{2}{c}{Attempts, $T$ ($n$)} \\
\cmidrule(lr){7-8}\cmidrule(lr){9-10}
 & 2024 & 2025 & 2026 & All & Mean & $\geq 5$ & $\geq 6$ & $\geq 5$ & $\geq 6$ \\
\midrule
Untutored & 63.3 & 60.0 & 58.3 & 60.6 & 0.97 & 15.6 & 11.1 & 41 & 20 \\
Untutored-Correct & \textcolor{blue}{65.8} & 57.1 & 56.7 & 59.9 & 1.04 & 13.3 & 11.1 & 48 & 29 \\
SPLICE-NF & 65.0 & 59.6 & 60.0 & 61.5 & \textbf{1.21} & 20.0 & 11.1 & 62 & 25 \\
SPLICE & \textbf{67.9} & \textbf{64.6} & \textcolor{blue}{60.8} & \textbf{64.4} & 1.15 & \textbf{24.4} & 13.3 & \textbf{70} & 41 \\
SPLICE-HC & 65.0 & \textbf{64.6} & \textbf{62.9} & \textcolor{blue}{64.2} & 1.14 & 20.0 & \textbf{17.8} & 69 & \textbf{45} \\
\bottomrule
\end{tabular}
\end{table}

\textbf{Main performance.} On AIME (Table~\ref{tab:results-2}), filtered SPLICE improves overall accuracy from 60.6\% to 64.4\% (64.2\% for SPLICE-HC); SPLICE-NF and Untutored-Correct remain near baseline. On proofs, every SPLICE variant raises the mean score, SPLICE by $+0.17$ and SPLICE-NF by $+0.24$, but only the filtered datasets improve its top: attempts scoring $\geq6$ double, from 20 to 41 and 45, and SPLICE-HC has existence at 6 of 17.8\% against the baseline's 11.1\%.\footnote{Paired over problems, 95\% bootstrap intervals, SPLICE / SPLICE-HC / SPLICE-NF against Untutored: AIME $+3.9$ [$+1.0$, $+6.9$], $+3.6$ [$+0.7$, $+6.7$], $+1.0$ [$-2.4$, $+4.3$] points; mean proof score $+0.17$ [$+0.09$, $+0.27$], $+0.16$ [$+0.07$, $+0.26$], $+0.24$ [$+0.14$, $+0.35$]; rate of attempts scoring $\geq5$, $+2.0$ [$+0.6$, $+3.8$], $+1.9$ [$+0.3$, $+4.1$], $+1.5$ [$+0.1$, $+3.3$] points. Against Untutored-Correct: AIME $+4.6$, $+4.3$, $+1.7$, the last not clear of zero; mean score $+0.11$ [$-0.01$, $+0.23$], $+0.10$ [$-0.02$, $+0.22$], $+0.17$ [$+0.09$, $+0.27$]; rate $\geq5$, $+1.5$, $+1.5$, $+1.0$, all three clear of zero.}

\textbf{Where the gain sits.}\label{sec:gain} Table~\ref{tab:results-5} shows that the best-of-$k$ gap grows with sampling: relative to Untutored, SPLICE gains $+0.21$ at $k=4$ and $+0.31$ at $k=32$, spread over 27 of the 45 problems. The gain is clear of zero through $k=8$ ($+0.20$, CI $+0.02$ to $+0.39$) and no longer resolved at $k=16$ (CI $-0.03$ to $+0.44$), since the best of many samples is an extreme statistic; SPLICE-NF's gain holds at every $k$. At thresholds 6 and above, filtering is decisive: SPLICE-NF sits within two points of the baseline curve while the filtered datasets lead it by three to six. Appendix~\ref{app:scores} gives the full score distribution.

\begin{table}[t]
\caption{Expected best-of-$k$ score, and pass@$k$ (\%), the probability that at least one of $k$ samples scores $\geq T$; both unbiased estimators over each problem's 32 samples, averaged over problems. pass@32 is the existence columns of Table~\ref{tab:results-2}, and pass@1 the share of attempts in Appendix~\ref{app:scores}.}
\label{tab:results-5}
\begin{center}
\footnotesize\setlength{\tabcolsep}{2.6pt}
\begin{tabular}{@{}lcccc cc cc cc cc@{}}
\toprule
 & \multicolumn{4}{c}{Best-of-$k$ score} & \multicolumn{8}{c}{pass@$k$ (\%)} \\
\cmidrule(lr){2-5}\cmidrule(lr){6-13}
 & & & & & \multicolumn{2}{c}{$T=4$} & \multicolumn{2}{c}{$T=5$} & \multicolumn{2}{c}{$T=6$} & \multicolumn{2}{c}{$T>6$} \\
\cmidrule(lr){6-7}\cmidrule(lr){8-9}\cmidrule(lr){10-11}\cmidrule(lr){12-13}
$k$ & 4 & 8 & 16 & 32 & 4 & 16 & 4 & 16 & 4 & 16 & 4 & 16 \\
\midrule
Untutored & 1.69 & 2.03 & 2.33 & 2.60 & 13.9 & 20.9 & 7.4 & 12.2 & 4.2 & 8.7 & 4.0 & 8.5 \\
Untutored-Correct & 1.82 & 2.18 & 2.52 & 2.85 & 15.9 & 25.6 & 8.3 & 12.2 & 6.2 & 10.8 & 5.5 & 10.5 \\
SPLICE-NF & \textbf{1.96} & \textbf{2.30} & \textbf{2.61} & 2.87 & \textbf{19.2} & \textbf{28.5} & 10.3 & 16.4 & 5.5 & 10.4 & 5.0 & 10.0 \\
SPLICE & 1.90 & 2.23 & 2.56 & 2.91 & 16.9 & 24.2 & \textbf{11.8} & \textbf{18.9} & 7.9 & 12.2 & 7.4 & 12.1 \\
SPLICE-HC & 1.90 & 2.25 & 2.58 & \textbf{2.94} & 17.9 & 26.2 & 10.3 & 16.1 & \textbf{8.3} & \textbf{14.2} & \textbf{8.0} & \textbf{14.1} \\
\bottomrule
\end{tabular}
\end{center}
\end{table}

\textbf{Conditioning.}\label{sec:conditioning} The gains are not explained only by reaching a gradeable write-up more often. Conditioned on natural termination, each SPLICE-trained student scores above Untutored; conditioned on attempts already scoring $\geq4$, the filtered datasets retain their advantage. Force-closed SPLICE attempts can also be useful when the cut removes repetition after a solution has effectively been found. Detailed termination and conditioning statistics are in Appendix~\ref{app:scores}. The proof results also separate \emph{breadth} from \emph{depth}: SPLICE-NF often produces reasonable attempts and has the highest mean score, but filtering matters from score 5 upward.

\subsection{What filtering changed}
\label{sec:filtering}

The generation record predicts what each filter changes. Removing 121
generation-failure traces (SPLICE-NF $\to$ SPLICE) costs a little at low
scores but improves the top of the proof distribution: at
score $\geq6$, SPLICE has 41 attempts on 6 problems against SPLICE-NF's 25
attempts on 5 problems, and among attempts already scoring $\geq4$ the share reaching 6
rises from 19\% to 33\%. AIME likewise moves from
+1.0 to +3.9 points over Untutored. These are the dimensions expected to
suffer when the target contains compression artifacts or a leaked reasoning
block.

Removing a further 219 traces with maximum-level hinting but no progress
(SPLICE $\to$ SPLICE-HC) has a different effect. Natural termination
rises from 820 to 927 of 1,440 attempts and the share of force-closed
attempts that had entered a repetition falls from 80\% to 70\%, while the
top of the distribution moves little, 45 attempts scoring $\geq6$ on 8
problems against 41 on 6. Thus the two filters do not behave as
generic quality filters: generation-failure filtering improves the top of the
score distribution, whereas excessive-hint filtering primarily improves
termination. This directional agreement matters because the filters were
defined from generation records rather than downstream test labels: the defects
identified during trace construction predict the dimensions of student behavior
that change after those traces are removed.

Untutored-Correct is the rejection-sampling contrast, and on this problem
set it cannot be scaled: the 768 traces are every problem the demonstrator
solves unaided. The question is whether the hinted traces add to them,
and they do: SPLICE-NF, which contains all 768, gains $+0.17$ on the mean
proof score over Untutored-Correct (CI $+0.09$ to $+0.27$), and the
filtered datasets gain $+4.6$ and $+4.3$ points on AIME, where
Untutored-Correct scores 59.9\% against the baseline's 60.6\%. What the
correct traces alone recover is $+0.07$ on the mean and 29 attempts
scoring $\geq6$ against the baseline's 20, on the same five problems, and
the student often fails to terminate. SPLICE adds training signal exactly
where rejection sampling has none: after an unaided trajectory has ceased
to make constructive progress.
Across SPLICE variants, more hinting in training is also associated with more
self-hinting and repetition at evaluation (Appendix~\ref{app:filtering-details}).

\subsection{What the trained student stores}
\label{sec:stores}

Hint tokens are only 3.5 to 4.6\% of the training tokens, but six times less probable under the untrained student than the demonstrator's own text, so they carry a fifth to a quarter of the initial finetuning loss.
With no tutor at evaluation, SPLICE and SPLICE-HC emit self-hint passages on
30.5\% and 23.1\% of proof samples (22.2\% and 15.0\% on AIME), before any
detected repetition; students trained without hints emit none. The proof/AIME
emission ratio also tracks the corresponding training-set hint ratio, suggesting
that the behavior is context-conditional rather than a generic verbal tic. We test whether these passages mark decision points by scoring evaluation traces
under the trained student and its pre-finetuning checkpoint, separating the
self-hint, up to three following paragraphs, and other text. Table~\ref{tab:results-10}
reports per-token loss and fork density (tokens per thousand with predictive
entropy above 2 nats).

\begin{table}[t]
\caption{Per-token NLL and fork density (tokens per thousand with predictive entropy $>2$ nats) by trace region. ``Following'' is up to three paragraphs after a self-hint; text after detected repetition is excluded.}
\label{tab:results-10}
\begin{center}
\footnotesize\setlength{\tabcolsep}{3pt}
\begin{tabularx}{\linewidth}{@{}l*{5}{>{\centering\arraybackslash}X}@{}}
\toprule
 & Region & NLL base & NLL trained & Relative drop & Fork density base $\to$ trained  \\
\midrule
Untutored & All text & 0.330 & 0.121 & 63\% & 6.8 $\to$ 2.9  \\
SPLICE & Other text & 0.399 & 0.133 & 67\% & 7.7 $\to$ 3.6  \\
 & Self-hint passage & 0.258 & 0.059 & 77\% & 10.7 $\to$ \textbf{12.5}  \\
 & Following text & 0.521 & 0.182 & 65\% & 17.3 $\to$ \textbf{35.0}  \\
SPLICE-HC & Other text & 0.404 & 0.136 & 66\% & 7.7 $\to$ 3.9  \\
 & Self-hint passage & 0.268 & 0.064 & 76\% & 11.7 $\to$ \textbf{14.4}  \\
 & Following text & 0.579 & 0.203 & 65\% & 18.0 $\to$ \textbf{37.6}  \\
\bottomrule
\end{tabularx}
\end{center}
\end{table}

Finetuning's default effect is confidence: in ordinary text, loss and fork
density fall. Around self-hints the pattern reverses selectively. Fork density
rises inside the passage and roughly doubles immediately after it, whereas the
passage itself becomes highly predictable (about 0.06 nats/token). Tutor-written
hint text remains much less probable than the student's own self-hints. The
pattern is therefore consistent with a stereotyped, student-generated cue that
marks an unresolved branch, followed by increased exploration, rather than
memorization of tutor wording. This analysis does not identify self-hinting as the cause of the score gain.
Pooled self-hinting attempts score half a point higher, but within-problem
comparisons remove most of that advantage ($+0.16$ for SPLICE, 95\% CI $+0.01$
to $+0.34$; $+0.12$ and $+0.08$ for SPLICE-NF and SPLICE-HC), indicating a
composition effect: self-hints occur most often on problems near the
student's capability boundary. Detailed
comparisons are in Appendix~\ref{app:scores}.

\section{Limitations and open problems}

The workflow's limits are set by what it does not know. The 2,500
problems were chosen without regard to where the demonstrator's ability
ends: some need no help, others stay unsolved at the highest hint level,
and every tutor call spent on either is wasted or worse, since the
second kind produces the excessive-hinting traces; predicting where a
hint can change the outcome would remove that failure at the source. The tutor is stronger than both the student
and the demonstrator, and we have not measured how much weaker it could
be. Its reasoning traces are not exposed, so distilling from the tutor is
not an available baseline; its final solutions carry no reasoning, and
finetuning on them is ordinary solution-only training, not a competing
recipe for reasoning traces.

The evidence has limits of its own. The entropy and self-hint analyses
show that SPLICE training produces a branching behavior; they do not
show that the behavior produces the score gain, and the test that
would, finetuning with the hint tokens masked from the loss, we have not
run. Our academic compute budget limited evaluation to AIME and 45 proof
problems; the intervals in Section~\ref{sec:results} reflect that, and a
larger proof set is needed before differences at a single threshold can
be read with confidence.

\section{Conclusion}

SPLICE moves verification and repair out of reinforcement learning and into offline data generation. By truncating failed continuations and inserting adaptively strengthened, restyled hints, it converts demonstrator failures into supervised traces unavailable to ordinary distillation or rejection sampling.

With student and demonstrator fixed, filtered SPLICE improves olympiad-proof grading and AIME accuracy. Filtering generation failures recovers the top of the proof-score distribution; removing excessive hinting improves termination. At test time, the student also emits self-hint-like branch cues at difficult points and explores more strongly afterward, showing that guidance in training traces can transfer a branching pattern while keeping the tutor out of inference.

\bibliography{references}
\bibliographystyle{iclr2027_conference}

\appendix
\section{Related work}
\label{app:related}

Mathematical reasoning in language models is measured and largely
trained on problems with a checkable final answer \citep{cobbe2021gsm8k,
hendrycks2021measuringmathematicalproblemsolving, he2024olympiadbench,
gao2024omnimathuniversalolympiadlevel, huang2024olympicarena}, a target
that grading model-written proofs showed does not carry over
\citep{mahdavi2025brains, petrov2025prooforbluff}. Proof writing
nonetheless advanced quickly: in 2025 frontier systems reached
gold-medal scores on the IMO problems \citep{huang2025gemini25imo,
LuongLockhart2025GeminiDeepThinkIMOGold, wei-openai-imo-2025-proofs},
and models have since contributed verified steps on open problems
\citep{bubeck2025earlyscience, tzachristas-etal-2026-open}. What has not
followed is a recipe without a heavy precondition: distilling from a
model that already writes proofs is sample efficient
\citep{muennighoff2025s1simpletesttimescaling, ye2025limoreasoning}, but
it requires such a model. The work below concerns proofs in natural
language, whose correctness no answer parser, test suite, or proof
checker can decide, grouped by what each line puts in place of that
missing check.

\textbf{Measuring proof quality.} Expert grading does not scale to an
experiment, so proof-level evaluation relies on a model as grader
\citep{zheng2023judgingllmasajudgemtbenchchatbot,
li2024llmsasjudgescomprehensivesurveyllmbased}, calibrated against
expert scores on annotated proof collections
\citep{dekoninck2025openproofcorpus, balunovic_srimatharena_2025}.
IMO-Bench pairs an expert-graded benchmark with an automatic grader
correlating 0.93 to 0.96 with human grades
\citep{luong2025towardsrobustmath}, and \citet{ma2025reliablefinegrained}
find that a strong reasoning backbone given the reference solution and a
marking scheme, ensembled over several runs, reaches a mean absolute
error of 0.926 against expert scores. Both agree on what makes a grader
reliable: reference material in the context, and aggregation over
several judgments. We propose no protocol of our own and apply that of
\citet{mahdavi2025brains}, whose agreement with human graders is
reported there, unchanged to every model we compare.

\textbf{Training without an automatic check.} Reinforcement learning with
verifiable rewards depends on a programmatic check of the final answer
\citep{deepseekai2026deepseekr1incentivizingreasoningcapability,
shao2024deepseekmathpushinglimitsmathematical,
kimiteam2025kimik15scalingreinforcement, lambert2025tulu3pushingfrontiers}.
Where none exists, a model produces the reward: a judge scoring the
response \citep{bai2022constitutionalaiharmlessnessai, pmlr-v235-lee24t,
pmlr-v235-yuan24d}, a generative verifier that reasons before scoring
\citep{zhang2025generativeverifiersrewardmodeling}, or a process reward
model scoring intermediate steps \citep{lightman2023letsverifystepstep,
wang-etal-2024-math, havrilla2024glorewhenwhereimprove}. The grader's
errors then become part of the objective:
\citet{dang2026escapingcognitivewell} document proof-refinement loops
converging on a wrong solution that both the solver and the pipeline's
own grader judge essentially correct, and DeepSeekMath-V2
\citep{deepseekai2025deepseekmathv2} answers the same reliability problem
by training a proof verifier, meta-verifying its output, and rewarding a
generator for proofs that survive it.

\textbf{Hints as a substitute signal.} A reward is useless if no rollout
earns it: when a policy fails on every rollout of a problem,
group-relative methods see identical rewards and the problems worth
learning from produce no gradient. Work on this injects external text
into the rollouts, surveyed under the term hint by
\citet{zhang2026surveyhintbasedrlvr}, and differs mainly in how the hint
is chosen. POPE \citep{qu2026popelearningreasonhard} supplies prefixes of
an oracle solution during training and reports that behavior learned
under guidance transfers to unguided problems; KnowRL
\citep{yu2026knowrlminimalsufficientknowledge} searches for a minimal
sufficient set of knowledge points, arguing that longer guidance adds
redundancy; ADHint \citep{zhang2025adhintadaptivehints} sets the amount
of hinting from per-sample difficulty; HiLL
\citep{xia2026learninghintreinforcementlearning} trains a hinter jointly
with the reasoner while penalizing trajectories whose correctness
depends on their hints; and LatentRevise
\citep{guo2026latentreviselearningzerohit} repairs the prefix of a
failed rollout in embedding space and finetunes on the continuations.

\textbf{Learning from a stronger model.} Where no usable reward is
available, a dataset can be built from a stronger model instead: by
distillation on its outputs \citep{hinton2015distillingknowledgeneuralnetwork,
kim-rush-2016-sequence}, for reasoning by finetuning on its traces
\citep{ho-etal-2023-large, magister-etal-2023-teaching,
hsieh-etal-2023-distilling}, or by keeping only the attempts that reach
a correct answer and iterating
\citep{zelikman2022starbootstrappingreasoningreasoning,
dong2023raftrewardrankedfinetuning,
gulcehre2023reinforcedselftrainingrestlanguage,
singh2024humandatascalingselftraining,
yuan2023scalingrelationshiplearningmathematical,
hosseini2024vstartrainingverifiersselftaught}. Conditioning the
demonstrating model on information the learner will not have at test
time is the privileged-information setting of imitation learning
\citep{pmlr-v15-ross11a, pmlr-v100-chen20a}, applied to reasoning by
conditioning a teacher on the solution before distilling it. That
variant has a documented failure:
\citet{kaur2026rethinkingonpolicyselfdistillation} report that
privileged distillation costs models post-trained for extended reasoning
up to 17 percent relative accuracy on competition benchmarks, because
privileged context lowers fork rates at high-entropy positions and
penalizes the tokens that open a self-correction branch, while the same
treatment helps instruction-tuned models.
\citet{zhang2026anchoredresidualprivilegeddistillation} give a
compatible account, that a solution-conditioned target lies outside the
student's local support and rewards skipping steps, and separate a
locally reachable anchor from an oracle residual.

\textbf{Systems for olympiad proof generation.} The strongest reported
proof results come from test-time scaffolds of repeated sampling,
checking, and revision \citep{NEURIPS2023_91edff07, NEURIPS2023_1b44b878,
NEURIPS2023_271db992, wang2023selfconsistencyimproveschainthought}, the
shape behind the 2025 gold-medal results \citep{huang2025gemini25imo,
LuongLockhart2025GeminiDeepThinkIMOGold, wei-openai-imo-2025-proofs}. A
scaffold presupposes the capabilities it calls, and below that threshold
models asked to correct their own errors without external feedback do
not reliably improve \citep{huang2024largelanguagemodelsselfcorrect},
which motivates training the capability instead
\citep{kumar2024traininglanguagemodelsselfcorrect,
hamid2026spirallearningsearchaggregate}. Where the domain admits a
symbolic checker the training signal is available directly, as in
formal provers \citep{AlphaGeometryTrinh2024,
ren2025deepseekproverv2advancingformalmathematical, chen2025seedprover,
lin2025goedelproverv2} and in InternGeometry
\citep{zhao2026interngeometry}, which reaches average gold-medalist
performance on IMO geometry against a symbolic engine. For
natural-language proofs at small scale, QED-Nano
\citep{qu2026qednanoteachingtinymodel} post-trains a 4B model by
supervised finetuning on traces distilled from DeepSeekMath-V2,
reinforcement learning against rubric rewards, and a stage that
decomposes long proofs into summarize-and-refine cycles.

\textbf{Positioning.} SPLICE puts a reference solution in place of the
missing check, consulted by a queried model that reports whether the
trace has advanced and writes a hint when it has not. QED-Nano is the
closest system in student scale, task, and two-stage shape, but its
supervised stage distils the traces of a model that already writes
proofs, which it therefore presupposes exists and is open; we ask a
strong model for no solution at all. We compare in a distillation setup,
a 30B demonstrator and a 4B student, because repeating a frontier
post-training pipeline is beyond our budget and because holding the
demonstrator fixed across both arms isolates the effect of the editing.
POPE shares our resource and spends it differently: reference solutions
enter the student's own rollouts inside the reinforcement learning loop,
so the student trains in a hinted condition, which HiLL's
transfer-weighted reward exists to correct for. We spend the resource
once, during data generation, on the demonstrator, and the student never
sees a hinted prompt.

The failure of privileged distillation reported by
\citet{kaur2026rethinkingonpolicyselfdistillation} bears directly on us,
and two features of our setting separate us from it: our student is
instruction-tuned rather than post-trained for extended reasoning, the
regime in which they report privileged context helps, and the tutor
never authors the trace, so the supervision target remains the
demonstrator's own text apart from an inserted hint rather than a
solution-conditioned rewrite of it. Our experiments do not separate
which of the two matters. Two further findings informed the design.
\citet{zhang2026anchoredresidualprivilegeddistillation} argue that
solution-conditioned targets push a student toward steps it could not
have reached, which is the case for truncating at the last verified step
and inserting the smallest hint that restores progress rather than
rewriting the trace. \citet{chandra2025shapeofthought} show that traces
ending in a wrong answer can still improve a student, so our truncation
of failed continuations describes the pipeline and does not claim that
keeping them would be worse.

\section{Models behind the tutor stages}
\label{app:tutor-models}

The tutor's stages (Table~\ref{tab:signatures}, Appendix~\ref{app:signatures}) run on three members of the
Gemini 2.5 family, chosen by cost. Decomposition, acceptance, and
compression, which read the whole reference solution or the whole
trace, run on Gemini 2.5 Pro; assessment, hint writing, and restyling
run on Gemini 2.5 Flash; detection runs on Gemini 2.5 Flash-Lite.

\begin{table}[h]
\caption{Model behind each tutor stage.}
\label{tab:tutor-models}
\begin{center}
\small
\begin{tabular}{@{}ll@{}}
\toprule
Stage & Model  \\
\midrule
$\textsc{Decompose}$, $\textsc{Accept}$, $\textsc{Shorten}$ & Gemini 2.5 Pro  \\
$\textsc{Assess}$, $\textsc{Hint}$, $\textsc{Restyle}$ & Gemini 2.5 Flash  \\
$\textsc{Stopped}$ & Gemini 2.5 Flash-Lite  \\
\bottomrule
\end{tabular}
\end{center}
\end{table}

\section{Generation constants}
\label{app:constants}

Table~\ref{tab:constants} lists the constants of the SPLICE loop
(Algorithm~\ref{alg:splice-loop}) at the values used to generate every
dataset in this paper. The repetition detector of Section~\ref{sec:loop}
fires on a repeated span of at least 200 tokens. For the first 200 tokens
after a hint is spliced in (Section~\ref{sec:following}), decoding
suppresses the tokens that open or close a reasoning block and the
following pivot words, in either case: \emph{wait, but, however,
alternatively, actually, hold, maybe, instead, no, oops, ah}. The demonstrator samples with its model card's recommended settings (temperature 0.6, top-\(p\) 0.95, top-\(k\) 20) and the tutor's Gemini~2.5~Pro stages at temperature 0.1; \(\textsc{Shorten}\) targets 80\% of the trace's length; \(\textsc{Stopped}\) is consulted after two consecutive chunks of fewer than 30 tokens or a chunk that closes the block with a tool-call or reasoning tag the demonstrator was not trained to emit.

\begin{table}[h]
\caption{Constants of the SPLICE loop.}
\label{tab:constants}
\begin{center}
\small
\begin{tabular}{@{}clr@{}}
\toprule
Symbol & Meaning & Value  \\
\midrule
$B$ & Iteration budget per problem & 15  \\
$L$ & Tokens per generation chunk & 16{,}000  \\
$L_{\mathrm{tot}}$ & Maximum trace length in tokens & 65{,}000  \\
$L_{\max}$ & Trace length at which $\textsc{Shorten}$ runs & 50{,}000  \\
$N$ & Acceptance samples that must all agree & 5  \\
\bottomrule
\end{tabular}
\end{center}
\end{table}

\section{Stage signatures}
\label{app:signatures}

Table~\ref{tab:signatures} collects the signature of every stage of
Algorithm~\ref{alg:splice-loop}, with the object each returns; Appendix~\ref{app:prompts} reproduces the prompt behind each.

\begin{table}[h]
\caption{The stages of trace generation. Subscripts give the model that runs each stage; the tutor's stages run on models of different cost, listed in Appendix~\ref{app:tutor-models}.}
\label{tab:signatures}
\begin{center}
\small
\begin{tabular}{@{}l>{\raggedright\arraybackslash}p{0.36\linewidth}>{\raggedright\arraybackslash}p{0.44\linewidth}@{}}
\toprule
Stage & Signature & Returns  \\
\midrule
Generation & $\textsc{Gen}_\demo(p, c, a)$ & The next chunk $x$  \\
Decomposition & $\textsc{Decompose}_{\tutor}(p, r)$ & $S = (s_1,\dots,s_m)$, the reference solution as ordered substeps under two to five main steps  \\
Acceptance & $\textsc{Accept}_{\tutor}(p, r, S, a)$ & A verdict in $\{0,1\}$ and a list $E$ of located errors  \\
Assessment & $\textsc{Assess}_{\tutor}(p, S, c, a, E)$ & Statuses $\sigma$, pointer $k$, cut index $u$  \\
Hint & $\textsc{Hint}_{\tutor}(p, r, S, c_{1:u}, a, \sigma, k, \ell, E)$ & The hint content $h_0$  \\
Restyling & $\textsc{Restyle}_{\tutor}(p, h_0, c_{1:u}, \Pi)$ & The hint $h$ as it enters the trace, given style patterns $\Pi$  \\
Compression & $\textsc{Shorten}_{\tutor}(c)$ & The same reasoning, shortened  \\
Detection & $\textsc{Stopped}_{\tutor}(p, c)$ & Whether a finished solution already sits in $c$  \\
\bottomrule
\end{tabular}
\end{center}
\end{table}

\section{Score distribution}
\label{app:scores}

Table~\ref{tab:results-4} gives the whole score distribution under the
aggregate grading protocol of Section~\ref{sec:proof}. The unfiltered SPLICE
traces shift the largest share of attempts away from zero, into partial
proofs and every band up to 6, but not past it; the filtered datasets shift
a smaller share but further up the scale, SPLICE-HC leading at 6 and above.
Untutored-Correct sits between the baseline and the SPLICE datasets in
every band.

\begin{table}[h]
\caption{Score distribution, as the percentage of attempts whose aggregate score falls in each band, the percentage scoring $\geq$ each threshold, and the number of attempts scoring $\geq 5$ and $\geq 6$.}
\label{tab:results-4}
\begin{center}
\footnotesize\setlength{\tabcolsep}{3pt}
\begin{tabularx}{\linewidth}{@{}l*{5}{>{\centering\arraybackslash}X}@{}}
\toprule
Score & Untutored & Untutored-Correct & SPLICE-NF & SPLICE & SPLICE-HC  \\
\midrule
$[0,1)$ & 69.7 & 66.9 & \textbf{62.8} & 66.5 & 67.3  \\
$[1,3)$ & 18.0 & 19.9 & \textbf{20.8} & 18.6 & 18.0  \\
$[3,4)$ & 6.0 & 6.0 & \textbf{7.3} & 6.2 & 6.0  \\
$[4,5)$ & 3.5 & 3.8 & \textbf{4.7} & 3.8 & 3.9  \\
$[5,6)$ & 1.5 & 1.3 & \textbf{2.6} & 2.0 & 1.7  \\
$[6,7]$ & 1.4 & 2.0 & 1.7 & 2.8 & \textbf{3.1}  \\
$\geq 3$ & 12.3 & 13.2 & \textbf{16.3} & 14.9 & 14.7  \\
$\geq 4$ & 6.3 & 7.2 & \textbf{9.0} & 8.7 & 8.7  \\
$\geq 5$ & 2.8 & 3.3 & 4.3 & \textbf{4.9} & 4.8  \\
$\geq 6$ & 1.4 & 2.0 & 1.7 & 2.8 & \textbf{3.1}  \\
\midrule
$n \geq 5$ & 41 & 48 & 62 & \textbf{70} & 69  \\
$n \geq 6$ & 20 & 29 & 25 & 41 & \textbf{45}  \\
\bottomrule
\end{tabularx}
\end{center}
\end{table}

\subsection{Termination and conditioning details}
\label{app:conditioning-details}
\begin{table}[t]
\caption{Left: attempts that finished naturally against attempts that were force-closed, with the count, mean score and number scoring $\geq 6$ of each, the share of force-closed attempts that had entered a repetition, and the mean score of force-closed attempts by where the cut fell (Untutored-Correct's 272 tool-call closures count as last-paragraph cuts). Right: mean score and share scoring $\geq 6$ among attempts already scoring $\geq 4$, and among those scoring $\geq 5$.}
\label{tab:results-9}
\begin{center}
\footnotesize\setlength{\tabcolsep}{2.4pt}
\begin{tabular}{@{}l rrr@{\hskip 6pt} rrrr@{\hskip 6pt} rr@{\hskip 6pt} rr@{\hskip 6pt} rr@{}}
\toprule
 & \multicolumn{3}{c}{Natural} & \multicolumn{4}{c}{Force-closed} & \multicolumn{2}{c}{Cut at} & \multicolumn{2}{c}{Given $\geq 4$} & \multicolumn{2}{c}{Given $\geq 5$} \\
\cmidrule(lr){2-4}\cmidrule(lr){5-8}\cmidrule(lr){9-10}\cmidrule(lr){11-12}\cmidrule(lr){13-14}
 & $n$ & mean & $\geq 6$ & $n$ & rep. & mean & $\geq 6$ & repetition & last par. & mean & $\geq 6$ & mean & $\geq 6$ \\
\midrule
Untutored & 1,268 & 0.99 & 20 & 171 & 51\% & 0.87 & 0 & --- & \textbf{0.87} & 5.12 & 22\% & 5.95 & 49\% \\
Untutored-Correct & 827 & 1.18 & 24 & 612 & 55\% & 0.84 & 5 & 1.01 & 0.77 & 5.17 & 28\% & 6.10 & 60\% \\
SPLICE-NF & 843 & \textbf{1.27} & 16 & 597 & 87\% & 1.14 & 9 & 1.20 & 0.69 & 5.10 & 19\% & 5.85 & 40\% \\
SPLICE & 820 & 1.12 & 17 & 620 & 80\% & \textbf{1.18} & \textbf{24} & \textbf{1.46} & 0.81 & 5.36 & 33\% & 6.13 & 59\% \\
SPLICE-HC & 927 & 1.16 & \textbf{28} & 513 & 70\% & 1.10 & 17 & 1.42 & 0.82 & \textbf{5.39} & \textbf{36\%} & \textbf{6.21} & \textbf{65\%} \\
\bottomrule
\end{tabular}
\end{center}
\end{table}

The termination split rules out a simple explanation in which the gains
come only from reaching a gradeable write-up more often. Conditioned on
finishing naturally, every student trained on SPLICE traces scores above
the Untutored baseline per finished attempt. Force-closed attempts also
contribute. For SPLICE they score above its natural finishes and carry
24 of its 41 attempts scoring $\geq 6$. The location of the cut helps
explain this: an attempt cut at the onset of a repetition scores
1.20--1.46 across the SPLICE students, above their natural finishes,
because it can already contain the substantive solution and differ from
a finished attempt mainly in stopping; an attempt cut at its last
paragraph scores only 0.69--0.82. Untutored-Correct's force-closed
attempts score 0.84; 272 of them emitted a tool-call token in place of
the end-of-reasoning marker and then nothing, so the forced closure was
effectively the only write-up.

Conditioning on attempts that have already reached a reasonable score
gives the same qualitative picture. Among attempts scoring at least 4,
the filtered datasets retain their advantage, most clearly for
SPLICE-HC: its reasonable attempts average 5.39, with 36\% reaching 6,
and 65\% of its attempts already scoring at least 5 reach 6.
SPLICE-NF's reasonable attempts are the weakest among the SPLICE-trained
students, averaging 5.10 with 19\% reaching 6. Thus the filtered models'
advantage is not merely an increase in the number of nontrivial
attempts; it also appears in how far already-promising attempts are
carried toward complete proofs.

\subsection{Self-hint attempt comparisons}
\label{app:selfhint-details}

Table~\ref{tab:results-11} asks a different question from the aggregate
self-hint rates: on the same problem, do attempts containing a self-hint
passage score higher than attempts without one? Pooled across problems,
a self-hinting SPLICE attempt scores about half a point higher and
reaches score 4 about twice as often. That pooled comparison is
confounded by problem composition, however. Within problem, most of the
advantage closes: $+0.16$ for SPLICE (28 problems won, 16 lost; 95\%
bootstrap CI $+0.01$ to $+0.34$), $+0.12$ for SPLICE-NF and $+0.08$ for
SPLICE-HC (20 won, 25 lost). The pooled gap is therefore primarily a
composition effect---self-hints occur disproportionately on
mid-difficulty problems near the student's capability boundary---and on
those same problems attempts containing a self-hint score only slightly
above attempts without one. Among attempts already scoring at least 4,
the number of problems containing both kinds of attempt is only 8--12, so
the corresponding differences are too small to support a strong
conclusion.

\begin{table}[t]
\caption{Attempts with at least one self-hint passage (before any detected repetition) against attempts with none. Pooled columns mix problems; the within-problem columns compare the two kinds of attempt on each of the 45 problems where both occur and report the mean difference and the count of problems on which the self-hinting attempts score higher, equal, or lower.}
\label{tab:results-11}
\begin{center}
\footnotesize\setlength{\tabcolsep}{3pt}
\begin{tabularx}{\linewidth}{@{}l*{5}{>{\centering\arraybackslash}X}@{}}
\toprule
 & Attempts with self-hint & Mean score, with / without & $P(\geq 4)$, with / without & Within problem, all attempts & Within problem, attempts $\geq 4$  \\
\midrule
SPLICE & 439 & 1.48 / 1.00 & 14.1\% / 6.3\% & +0.16 (28 / 1 / 16) & $-$0.11 (4 / 0 / 4, $n = 8$)  \\
SPLICE-HC & 332 & 1.38 / 1.07 & 13.3\% / 7.3\% & +0.08 (20 / 0 / 25) & +0.06 (4 / 0 / 6, $n = 10$)  \\
SPLICE-NF & 440 & 1.30 / 1.17 & 12.0\% / 7.7\% & +0.12 (26 / 0 / 19) & +0.07 (8 / 0 / 4, $n = 12$)  \\
\bottomrule
\end{tabularx}
\end{center}
\end{table}

These comparisons sharpen the interpretation of self-hinting. The
trained student has learned a recognizable behavior that is absent from
the no-hint controls and is deployed preferentially on problems where
the model is partially capable, but these observational comparisons do
not establish that self-hinting causes the score improvement. Training
could produce both the performance gain and the self-hint behavior, and
the text following a self-hint differs from ordinary text in content as
well as position. A direct intervention---for example, finetuning with
hint tokens masked from the loss and testing whether the gains
survive---would be needed to establish a causal role.

\section{Filtering, hint exposure, and diversity analyses}
\label{app:filtering-details}

\subsection{Detailed filtering comparisons}

\paragraph{Generation failures: SPLICE-NF to SPLICE.}
Removing 121 traces primarily changes the top of the score distribution.
SPLICE-NF leads SPLICE at every score below 5 and in expected
best-of-\(k\) up to \(k = 16\), but at scores 5 and 6 it falls behind:
it produces 25 attempts scoring \(\geq 6\) against SPLICE's 41 and has
the weakest already-reasonable attempts of the trained students. On AIME, the same removal changes the
gain over Untutored from +1.0 percentage point, within the reported
confidence interval, to +3.9 points, outside it. This directional pattern
is consistent with the trace-construction diagnosis: compression
artifacts and leaked reasoning blocks are
most plausibly harmful to high-quality completions rather than to the
mere production of partial work. Because these filters were defined
from generation records rather than downstream test labels, this
agreement is diagnostic rather than a post-hoc selection on evaluation
scores.

\paragraph{Excessive hinting: SPLICE to SPLICE-HC.}
Removing 219 traces in which hint level 4 was reached without progress
changes a different axis. SPLICE-HC finishes naturally on 927 proof
attempts against SPLICE's 820; among force-closed attempts, 70\% rather
than 80\% had entered a repetition. SPLICE-HC also produces 45 attempts
scoring \(\geq 6\) on 8 problems, compared with SPLICE's 41 on 6, and
has the strongest already-reasonable attempts, while SPLICE leads by a
fraction of a point at every threshold below 6.
Thus the two filters have distinct empirical signatures:
generation-failure filtering primarily improves the top of the score
distribution, whereas excessive-hint filtering primarily improves
termination and repetition behavior.

\paragraph{Repair versus selection: Untutored-Correct.}
Rejection sampling retains the 768 traces on which the demonstrator
succeeded unaided. Its student remains close to Untutored on AIME and
from score 5 upward, gaining 0.07 on the mean, and has an attempt scoring
\(\geq 6\) on five problems, as does the baseline. Its evaluation-time termination behavior
is distinctive: 76\% of attempts exit through a tool-call token rather
than the end-of-reasoning marker, 19\% emit nothing afterward, and 42\%
require force-closing; the force-closed attempts average 0.84. Thus the
available set of clean unaided successes recovers part of the mean gain
of the repaired traces but none of their gain at score \(\geq 6\) or on
AIME. The set cannot be enlarged: the 768 traces are every problem the
demonstrator solves unaided on this problem set.

\subsection{Hint exposure across datasets}
The amount of hinting in training rises from none in Untutored-Correct
through SPLICE-HC and SPLICE to SPLICE-NF. At evaluation, counting only
text before any detected repetition, the share of proof-grid samples
containing at least one self-hint passage is 0.0\%, 23.7\%, 30.6\%, and
30.6\%, respectively. Among force-closed attempts, the share that had
entered a repetition is 55\%, 70\%, 80\%, and 87\%; the corresponding
numbers of attempts scoring \(\geq 6\) are 29, 45, 41, and 25. Across these
datasets, greater hint exposure is therefore associated with more
self-hinting and more repetition at evaluation, while filtering the
highest-hint/no-progress traces improves termination. These are
cross-dataset associations, not an identified dose-response effect.
Untutored-Correct, which saw no hints, emits no detected self-hint in
1,440 samples, providing a negative control for the detector used in
Section~\ref{sec:stores}.

\subsection{Detailed self-hint likelihood and fork-density analysis}
\label{app:selfhint-likelihood}
The main text reports that finetuning has a generic confidence effect:
for ordinary text, per-token loss and fork density fall. We additionally
examined three qualitative signatures before looking at the regional
measurements. If a self-hint marks a \emph{decision point}, uncertainty
should remain elevated in or immediately after the cue even as ordinary
text becomes more predictable. If it is primarily \emph{memorized tutor
text}, the trained model should instead become unusually confident on
the hint wording itself without a corresponding rise in uncertainty
afterward. If it is \emph{inert stylistic text}, its local uncertainty
profile should resemble ordinary text.

The observed pattern is closest to the first account, with an important
qualification. Under finetuning, fork density falls in ordinary text but
rises inside self-hint passages, from 10.7 to 12.5 per thousand tokens
for SPLICE and from 11.7 to 14.4 for SPLICE-HC. In the text immediately
following a self-hint it rises further, to roughly 35 and 38 per
thousand, the highest of the measured regions. At the same time, the
self-hint phrase itself becomes highly predictable under the trained
student (about 0.06 nats/token in the regional analysis reported in
Table~\ref{tab:results-10}). The cue can therefore be stereotyped even
while the continuation that follows it becomes locally more exploratory.

The wording also does not appear to be copied directly from the tutor.
When scored under the trained student, tutor-written hint text costs
about 1.771 nats/token, whereas the student's own detected self-hints
cost about 0.26 nats/token, lower than its ordinary reasoning text.
Taken together, these measurements are consistent with the student
learning a cue in its own distribution that marks a difficult branch,
rather than memorizing the tutor's surface form. They do not establish
that this behavior causes the performance gain; the within-problem
comparisons in Appendix~\ref{app:selfhint-details} are deliberately more
cautious on that point.

\subsection{Approach coverage}
\label{sec:coverage}
Reference-conditioned hinting could in principle collapse solutions
toward the reference approach. Every graded attempt carries the
reference cluster it was matched to, computed per problem and shared
across students. The filtered datasets leave coverage unchanged: SPLICE
and SPLICE-HC attempt 3.38 and 3.42 distinct approaches per problem
against the baseline's 3.40 (paired differences within \(\pm 0.1\), 95\%
bootstrap CI about \(\pm 0.3\)), and the number of approaches with a
sample scoring \(\geq 3\) is 31 to 35 for every student. The unfiltered
traces narrow it: SPLICE-NF attempts 3.09 approaches per problem,
\(-0.31\) against Untutored (CI \(-0.69\) to \(+0.07\)) and \(-0.40\)
against Untutored-Correct (CI \(-0.71\) to \(-0.09\)), losing on 20
problems and gaining on 8. On AIME the same analysis is null in every
slice, about 1.4 observed approaches per problem against 3.4 available,
because AIME admits essentially one method per problem. The measure is
the reference-derived clustering, not semantic diversity in general, and
an effect of a few tenths of a cluster is not excluded for the filtered
datasets.

\section{Grading protocol}
\label{app:grading}

Proofs are graded with the protocol of \citet{mahdavi2025brains}. Per
problem, the reference solutions are gated for completeness and
clustered into approaches; each sample is matched to the cluster it
follows, a rubric is authored for that cluster, and the judge scores the
final write-up from 0 to 7 conditioned on the rubric. Clusters and
rubrics are generated once per problem and shared by every judge and
every student, so no student is graded against a rubric the others did
not face. Three judges take part: Nemotron-3.5-Lightning-30B-A3B,
sampled repeatedly at temperature, and Gemini 2.5 Pro and Gemini 3.1
Pro, one draw each. The reported score is the equal-weighted mean of the
three, Nemotron entering as the mean of its five draws.

Table~\ref{tab:judges} validates the instrument on 90 human-graded
proofs. A single Nemotron draw is not usable, with roughly 8 of the 90
cells far from the human grade; repeated draws alone recover the quality
of the published protocol, and combining decorrelated judges
(Nemotron--Gemini \(r \approx 0.8\)) is what passes 0.8. Weights fitted
on the 90 proofs do no better under leave-one-out (\(r = 0.789\)), so we
keep equal weights. The two Gemini versions differ systematically: 2.5
grades 0.34 higher than 3.1 on the same write-ups and rubrics, adding
partial credit to work 3.1 scores at zero and withholding sevens, and
the human grades side with 3.1 on every metric; a comparison that mixes
judges across students is therefore not valid, and every number in the
paper uses the same judges for every student.

\begin{table}[h]
\caption{Agreement with the 90 human-graded proofs (Pearson $r$).}
\label{tab:judges}
\begin{center}
\small
\begin{tabular}{@{}lc@{}}
\toprule
Judge & $r$ \\
\midrule
Nemotron, single draw & 0.563 \\
Nemotron, mean of five draws & 0.726 \\
Gemini 2.5 Pro & 0.742 \\
Gemini 3.1 Pro & 0.766 \\
Published protocol \citep{mahdavi2025brains} & 0.730 \\
Equal-weighted mean of the three & \textbf{0.803} \\
\bottomrule
\end{tabular}
\end{center}
\end{table}

\subsection{Judge failure on empty write-ups}
\label{app:judge-failure}

Force-closing (Section~\ref{sec:protocol}) appends a closure to an
unfinished attempt and lets the student write the final solution. For
some attempts, almost all of them from the Untutored-Correct student,
that closure produces no mathematics: the student re-emits the prompt,
with a spurious \texttt{user} turn between copies, for tens of thousands
of characters. Both Gemini judges score such a write-up 0. Nemotron
awards it 7 on every one of its five draws, with a rationale that
describes a complete proof (Table~\ref{tab:empty-writeups}); the
five-draw mean does not protect against a failure the judge repeats
consistently. Averaged with the two zeros, the attempt still receives a
score above 2, and because these attempts land on problems the student
otherwise fails, they inflate its best-of-\(k\) scores well beyond what
its other attempts support.

The calibration set of Table~\ref{tab:judges} contains no empty
write-ups, so the equal-weighted mean was never tested on this case. We
therefore add one rule to the protocol: when Nemotron is the highest of
the three judges and the three scores span four points or more, the
attempt takes Gemini 3.1's score alone. The rule fires on 53 of the
7,199 graded attempts (5, 25, 12, 7 and 4 for Untutored,
Untutored-Correct, SPLICE-NF, SPLICE and SPLICE-HC). On the calibration
set it fires on 2 of 90 proofs and lowers agreement with the human
grades from \(r = 0.802\) to \(0.770\), a cost we accept because the
calibration set lacks the failure the rule addresses; with the threshold
at five points the rule never fires on the calibration set and leaves
16 of Untutored-Correct's empty write-ups uncorrected. Existence and
pass@\(k\) are unaffected by the rule at every threshold; it changes
mean scores by at most 0.04 and the best-of-\(k\) columns of
Table~\ref{tab:results-5}.

\begin{table}[h]
\caption{Three empty write-ups and their scores. Length is the graded text; the Nemotron rationale is from its first draw.}
\label{tab:empty-writeups}
\begin{center}
\small
\begin{tabular}{@{}lrrrrp{0.42\linewidth}@{}}
\toprule
Student & Length & Nemotron & G2.5 & G3.1 & Nemotron rationale \\
\midrule
Untutored-Correct & 97,727 & 7.0 & 0 & 0 & ``Perfect solution executing all key strategic insights and sub-steps correctly, fully marking the orthocenter\ldots'' \\
Untutored-Correct & 145,397 & 7.0 & 0 & 0 & ``The solution provides a complete, rigorous synthetic geometry proof. All key strategic insights\ldots'' \\
Untutored-Correct & 102,788 & 7.0 & 4 & 0 & (no rationale returned) \\
\bottomrule
\end{tabular}
\end{center}
\end{table}

\section{Prompts of the SPLICE workflow}
\label{app:prompts}

The prompts below are the templates behind the stages of Algorithm~\ref{alg:splice-loop}, reproduced verbatim from the configuration that generated the datasets; braces mark the slots the orchestrator fills (the problem statement, the reference solution, the current trace, the hint level). Typographic characters are rendered in ASCII. \textsc{Gen} is the demonstrator's system prompt; every other stage is a tutor prompt, Appendix~\ref{app:tutor-models} lists the model behind each and Appendix~\ref{app:contracts} the inputs and outputs of each.

\(\textsc{Gen}_{\demo}\) is the only prompt the demonstrator sees; generation resumes from the retained prefix under this system prompt and the problem.

\promptbox{promptdemo}{$\textsc{Gen}_{\demo}$}{prompts/gen.txt}

\(\textsc{Decompose}_{\tutor}\) reads the problem and the reference solution \(r\) once per instance and returns the substep list \(S\) that every later stage grades against.

\promptbox{prompttutor}{$\textsc{Decompose}_{\tutor}$}{prompts/decompose.txt}

\(\textsc{Accept}_{\tutor}\) receives the problem, the reference solution and the final solution \(a\), never the trace, and returns a verdict with the list \(E\) of errors located in \(a\).

\promptbox{prompttutor}{$\textsc{Accept}_{\tutor}$}{prompts/accept.txt}

\(\textsc{Assess}_{\tutor}\) receives the trace \(c\) with numbered paragraphs, the substeps \(S\) and the reference, and returns the status map \(\sigma_t\), the pointer \(k_t\) and the last constructive paragraph \(u_t\).

\promptbox{prompttutor}{$\textsc{Assess}_{\tutor}$}{prompts/assess.txt}

\(\textsc{Hint}_{\tutor}\) receives the errors \(E\), the assessment, the target substep and the hint level \(\ell\), and returns the mathematical content of the hint at that level.

\promptbox{prompttutor}{$\textsc{Hint}_{\tutor}$}{prompts/hint.txt}

\(\textsc{Restyle}_{\tutor}\) receives the hint content, the retained prefix and the demonstrator's language patterns \(\Pi\), and returns the paragraphs that are spliced into the trace.

\promptbox{prompttutor}{$\textsc{Restyle}_{\tutor}$}{prompts/restyle.txt}

\(\textsc{Stopped}_{\tutor}\) receives a chunk and reports whether a finished solution is already sitting inside the reasoning block.

\promptbox{prompttutor}{$\textsc{Stopped}_{\tutor}$}{prompts/stopped.txt}

\(\textsc{Shorten}_{\tutor}\) receives the trace and a target factor when \(c\) exceeds \(L_{\max}\) tokens and returns the shortened trace; Section~\ref{sec:failure-modes} measures the damage it does.

\promptbox{prompttutor}{$\textsc{Shorten}_{\tutor}$}{prompts/shorten.txt}

\clearpage
\section{Input--output contracts of the stages}
\label{app:contracts}

Table~\ref{tab:contracts} states, for each stage of Algorithm~\ref{alg:splice-loop}, what the orchestrator places in the prompt of Appendix~\ref{app:prompts} and what it reads back. Three stages return structured JSON validated against a schema; the others return text that the orchestrator uses whole. The constants \(N\), \(T\), \(L\) and \(L_{\max}\) are those of Appendix~\ref{app:constants}.

\begin{table}[htbp]
\caption{What each stage receives and returns, and how the orchestrator uses the return.}
\label{tab:contracts}
\begin{center}
\scriptsize
\begin{tabularx}{\linewidth}{@{}>{\raggedright\arraybackslash}p{0.15\linewidth}>{\raggedright\arraybackslash}X>{\raggedright\arraybackslash}X@{}}
\toprule
Stage & Receives & Returns, and what the orchestrator does with it \\
\midrule
\(\textsc{Gen}_{\demo}\) & System prompt; the problem \(p\) as the user turn; on resumption, the retained prefix \(c_{1:u}\) with the spliced hint as the assistant turn. & A chunk of at most \(L\) tokens of free text. \(\textsc{Paragraphs}\) splits it at blank lines and appends the pieces to \(c\) as new steps; \(\textsc{Repeats}\) and \(\textsc{Stopped}\) are then run on the chunk. \\
\addlinespace
\(\textsc{Decompose}_{\tutor}\) & \(p\), the reference solution \(r\). & Free text: main steps, each with numbered substeps \(i.j\) (the list \(S\)). Passed verbatim as \texttt{solution\_analysis} to \(\textsc{Accept}\), \(\textsc{Assess}\) and \(\textsc{Hint}\); computed once per instance. \\
\addlinespace
\(\textsc{Accept}_{\tutor}\) & \(p\), \(r\), \(S\), and the final solution \(a\) alone (never the trace). & JSON \{\texttt{correctness}: bool; \texttt{errors}, \texttt{contradiction\_errors}: lists of \{\texttt{type}, \texttt{description}, \texttt{location}\}, the location a verbatim snippet of \(a\); \texttt{explanation}\}. Sampled \(N\) times; the trace is accepted only if every sample returns \texttt{correctness} true. Otherwise the error lists are formatted into the report \(E\) given to \(\textsc{Assess}\) and \(\textsc{Hint}\). Error types: proof by example, proposal without verification, inventing wrong facts, circular reasoning, trial and error, calculation mistake, missing closing token or hallucinated content, contradiction with the reference. \\
\addlinespace
\(\textsc{Assess}_{\tutor}\) & \(p\), \(S\), the trace \(c\) with its paragraphs labelled Step 1, 2, \ldots, the solution \(a\), and \(E\). & JSON \{\texttt{progress\_pointer}: \{\texttt{main\_step}, \texttt{substep}\}; \texttt{remaining\_work\_summary}; \texttt{last\_useful\_step\_index} (1-based, into the labelled steps); \texttt{last\_useful\_step\_reasoning}; \texttt{step\_status}: list of \{\texttt{step\_id}, \texttt{status} \(\in\) \{correct, incorrect, missing\}\}\}. The status list is \(\sigma_t\); the count of \emph{correct} entries is \(v_t\), compared with the running maximum \(V_{t-1}\) to decide stall or progress and move \(\ell\) down or up by one within \([0,4]\); \texttt{last\_useful\_step\_index} is \(u_t\), and the trace is cut to \(c_{1:u_t}\); the pointer gives \(k_t\). \\
\addlinespace
\(\textsc{Hint}_{\tutor}\) & \(p\), \(r\), \(S\), \(c_{1:u}\) labelled by step, \(a\), the formatted assessment, the level \(\ell\), and \(E\). & Free text: the content of the hint at level \(\ell\). Passed to \(\textsc{Restyle}\); recorded verbatim in the trace's generation record. \\
\addlinespace
\(\textsc{Restyle}_{\tutor}\) & \(p\), the hint content, \(c_{1:u}\), and the demonstrator's language patterns \(\Pi\) (a fixed list mined from its untutored traces). & Free text in the demonstrator's idiom. Split at blank lines; the pieces are appended to \(c_{1:u}\) as the spliced steps, and generation resumes after them. If the return is empty the raw hint is spliced instead. \\
\addlinespace
\(\textsc{Stopped}_{\tutor}\) & \(p\) and the chunk's paragraphs labelled Paragraph 1, 2, \ldots & JSON \{\texttt{has\_solution}: bool; \texttt{solution\_start\_\allowbreak paragraph\_index} (1-based); \texttt{reasoning}\}. If true, \(\textsc{SplitAtSolution}\) cuts \(c\) at that paragraph and the remainder becomes \(a\). \\
\addlinespace
\(\textsc{Shorten}_{\tutor}\) & The trace \(c\) and a target length (80\% of the input). & Free text: the shortened trace. Replaces \(c\) only if shorter than the input, after stripping any solution header the rewrite introduced; invoked when \(c\) exceeds \(L_{\max}\) tokens. \\
\bottomrule
\end{tabularx}
\end{center}
\end{table}

\clearpage
\section{Training and evaluation settings}
\label{app:training}

Table~\ref{tab:sft} gives the finetuning recipe. It is identical for the five datasets of Section~\ref{sec:datasets}; only the training file changes. Untutored, SPLICE and SPLICE-HC were trained on one 8-GPU node of the cluster and SPLICE-NF and Untutored-Correct on an 8-GPU node of a cloud provider, from the same pinned package set, so the recipe is the same to the package version. Table~\ref{tab:evalsettings} gives the evaluation settings shared by every student.

\begin{table}[htbp]
\caption{Supervised finetuning recipe, the same for every dataset.}
\label{tab:sft}
\begin{center}
\small
\begin{tabularx}{\linewidth}{@{}l>{\raggedright\arraybackslash}X@{}}
\toprule
Setting & Value \\
\midrule
Base model & Qwen3-4B-Instruct-2507 \\
Objective & next-token cross-entropy over the chat-formatted trace and solution \\
Precision, attention & bfloat16, FlashAttention-2 \\
Sequence length & 49,152 tokens, no truncation (the longest training rows are about 42,000 tokens) \\
Packing & sequences packed to 49,152 tokens; one pack per GPU per step \\
Global batch & 8 packs per optimizer step, about 410,000 tokens; no gradient accumulation \\
Epochs & 8 (about 600 optimizer steps per dataset) \\
Optimizer & AdamW, \(\beta = (0.9, 0.999)\), \(\epsilon = 10^{-8}\), weight decay 0.01 \\
Learning rate & \(10^{-5}\) peak, cosine decay to \(10^{-6}\), 20 warm-up steps (40 for Untutored) \\
Gradient clipping & global norm 1.0 \\
Seed & 42 (data shuffle and initialisation) \\
Checkpoint evaluated & the final one, after epoch 8 \\
Hardware & 8 \(\times\) H100, one node \\
\bottomrule
\end{tabularx}
\end{center}
\end{table}

\begin{table}[htbp]
\caption{Evaluation settings, the same for every student.}
\label{tab:evalsettings}
\begin{center}
\small
\begin{tabularx}{\linewidth}{@{}l>{\raggedright\arraybackslash}X@{}}
\toprule
Setting & Value \\
\midrule
Serving & vLLM, bfloat16, context 65,536 tokens \\
Sampling & temperature 0.7, top-\(p\) 0.8, top-\(k\) 20 (the base model's recommended values); at most 50,000 new tokens \\
Prompt & the problem alone, under the chat template, with the assistant turn opened by the reasoning-block marker so the student reasons before writing the solution \\
Proofs & 45 problems \(\times\) 32 samples; unfinished attempts force-closed as in Section~\ref{sec:protocol} \\
AIME & 90 problems \(\times\) 8 samples; an attempt without a boxed answer is wrong \\
Grading & Appendix~\ref{app:grading} \\
\bottomrule
\end{tabularx}
\end{center}
\end{table}

\end{document}